\documentclass[conference]{IEEEtran}
\makeatletter
\def\ps@headings{%
\def\@oddhead{\mbox{}\scriptsize\rightmark \hfil \thepage}%
\def\@evenhead{\scriptsize\thepage \hfil \leftmark\mbox{}}%
\def\@oddfoot{}%
\def\@evenfoot{}}
\makeatother
\usepackage{multirow}
\usepackage{amssymb,amsmath,makeidx}
\usepackage{amsthm}
\usepackage[left,modulo]{lineno}
\usepackage{bm}
\usepackage{psfrag}
\usepackage{cite} 
\usepackage{cite}
\usepackage{times}
\usepackage{graphicx}
\usepackage{amsmath,amssymb}
\usepackage{bbm}
\usepackage{graphicx,subfigure}
\usepackage[english]{babel}
\usepackage{dsfont}

\newif\ifnotes\notestrue

\def\hgr#1{}

\newcommand{\ben}{\begin{enumerate}}
\newcommand{\een}{\end{enumerate}}

\newcommand{\bc}{\begin{center}}
\newcommand{\ec}{\end{center}}

\newcommand{\bit}{\begin{itemize}}
\newcommand{\eit}{\end{itemize}}

\newcommand{\ds}{\displaystyle}
\newcommand{\beq}{\begin{equation}}
\newcommand{\eeq}{\end{equation}}

\newcommand{\ppuv}{p^+_{v,u}}
\newcommand{\pnuv}{p^-_{v,u}}

\newcommand{\wpuv}{w^+_{v,u}}
\newcommand{\wnuv}{w^-_{v,u}}

\newcommand{\wpvu}{w^+_{u,v}}
\newcommand{\wnvu}{w^-_{u,v}}

\newcommand{\lpu}{\lambda^+_{u}}
\newcommand{\lnnu}{\lambda^-_{u}}

\newcommand{\load}{\varrho}
\newcommand{\loadu}{\varrho_u}

\newcommand{\Tpu}{T^+_u}
\newcommand{\Tnu}{T^-_u}

\newcommand{\wre}{\mathbf{w}^{\rm{r}}}
\newcommand{\wi}{\mathbf{w}^{\rm{in}}}
\newcommand{\wo}{\mathbf{w}^{\rm{out}}}

\newcommand{\va}{\mathbf{a}}
\newcommand{\vb}{\mathbf{b}}

\newcommand{\x}{\mathbf{x}}
\newcommand{\y}{\mathbf{y}}

\newcommand{\vak}{\va^{(k)}}
\newcommand{\vbk}{\vb^{(k)}}

\newcommand{\Na}{N_{\rm{a}}}
\newcommand{\Nb}{N_{\rm{b}}}
\newcommand{\Nx}{N_{\rm{x}}}

\newcommand{\State}{\mathbf{S}}

\newcommand{\Prob}{\mathds{P}}

\newcommand{\R}{\mathds{R}}

\begin{document}
\title{Echo State Queueing Network: \\
a new reservoir computing learning tool}
%
%%%%%
%\author{\IEEEauthorblockN{~}
%\IEEEauthorblockA{~\\
%~\\
%~\\
%~}
%\and
%\IEEEauthorblockN{~}
%\IEEEauthorblockA{~\\
%~\\
%~\\
%~}
%}

\author{\IEEEauthorblockN{Sebasti\'an Basterrech}
\IEEEauthorblockA{INRIA-Rennes, France\\
Sebastian.Basterrech@inria.fr}
\and
\IEEEauthorblockN{Gerardo Rubino}
\IEEEauthorblockA{INRIA-Rennes, France\\
Gerardo.Rubino@inria.fr}
}


\maketitle

\begin{abstract}
%\boldmath
%

In the last decade, a new computational paradigm was introduced
in the field of Machine Learning, under the name of
\textit{Reservoir Computing} (RC). 
RC models are neural networks which a recurrent part
(the \textit{reservoir}) that \textit{does not participate}
in the learning process, and the rest of the system where no
recurrence (no neural circuit) occurs.
This approach has grown rapidly due to its success in solving
learning tasks and other computational applications.
Some success was also observed with another recently proposed
neural network designed using Queueing Theory, the Random Neural
Network (RandNN).
Both approaches have good properties and identified drawbacks.

In this paper, we propose a new RC model called
\textit{Echo State Queueing Network} (ESQN), where we use ideas
coming from RandNNs for the design of the reservoir.
ESQNs consist in ESNs where the reservoir has a new dynamics
inspired by recurrent RandNNs.
The paper positions ESQNs in the global Machine Learning area, and provides
examples of their use and performances.
We show on largely used benchmarks that ESQNs are very accurate tools,
and we illustrate how they compare with standard ESNs.

%This model can be used to solve a large variety of machine learning problems, specially  problems relate  networking, is a new tool which can be considered at the moment of solve to to apply 
%
%
\emph{Index Terms - Reservoir Computing, Echo State Network, Random Neural Network, Queueing Network, Machine Learning}\\

\end{abstract}

\section{Introduction}
\label{sec:Introduction}

Artificial Neural Networks (ANNs) are a class of computational models which have been proven to be very powerful as statistical learning tools to solve complicated engineering tasks as well as many theoretical issues.
%consumer networking technologies and applications
%
Several types of ANNs have been designed, some of them originating in the field of Machine Learning while others coming from biophysics and neuroscience.
%
%%%
%Definir SSN
%A specific type of ANNs is the Spiking Neuron Networks (SNN) class, in this network type the abstraction of the biological neuron is realized by an information processing unit named spiking neuron.
%
%The neurons send spikes 
%
%In contrast to traditional ANNs models (such as Multi-Layer Perceptron (MLP)), the SSN models the exact timing of their spikes among the neurons.
%
%The firing time representation can be a means for coding information.
%
%Another difference is that SSN models are for real--time computations on continuous spikes sequences which are originated from the environment (abstractions of  cortical microcircuits).\\
%
%
% streams of data. The spikes are sequences of action potentials of neurons that provide external inputs from the environment.
% The synapses representation is different and  has a model to precise representation of time firing 
%
% The essential difference between SNN and  is in the synapses representation.
%
% which assign meaning to the presence and timing of spikes  new notions are considered with respect the MLP models.
%that  Each individual spike is   and an activation function to determinate the output of each neuron.\\
%
The Random Neural Network (RandNN) proposed by E. Gelenbe in 1989~\cite{Gel89:RNNPosNeg}, is a mathematical object inspired by biological neuronal behavior which merges features of Spiking Neural Networks and Queueing Networks.
In the literature, actually two different interpretations of exactly the same mathematical model are proposed. 
One is a type of spiking neuron and the associated network which is called RandNNs. 
The other one is a new type of queue and networks of queues, respectively called G-queues and G-networks.
The RandNN is a connectionist model where spikes circulate among the interconnected neurons.
%
%represented by a direct point process 
A discrete state-space is used to represent the internal state (potential) of each neuron. 
The firing times of the spikes are modeled as Poisson processes.
The potential of each neuron is represented by a positive integer
that increases when a spike arrives or decreases after the neuron fires.
%a simple counter which counts negative $(-1)$ and positive (+1).
%
In order to use RandNNs in supervised learning problems,
a gradient descent algorithm has been described in~\cite{Gelenbe1993c},
and Quasi-Newton methods have been proposed in~\cite{Likas00, Baster09}.
Additionally, the function approximation properties of the model were studied in~\cite{Gel98:LearningAndApprox,GelMao04:Approx}.
The structure of the model leads to efficient numerical evaluation procedures, to
good performance in learning algorithms and to easy hardware implementations. Consequently, since its introduction the model has been applied in a variety of scientific fields.
Nevertheless, the RandNNs model suffers from limitations.
Some of them are related to the use of a feedforward topology
(see~\cite{Kocak11}).
The original acronym to refer the model was RNN. In this work to avoid a conflict of notation, we use RandNN for Random Neural Networks, due to the use of RNNs in Machine Learning literature for Recurrent Neural Networks.

%Other limitations comes from the recurrent architecture which affects in supervised learning.

Concerning models with recurrences (circuits) in their topologies,
they are recognized as powerful tools for a number of tasks in Machine Learning (both traditional ANNs and RandNNs). However,
they have a main limitation which comes from the difficulty in implementing efficient
training algorithms.
The main drawbacks related to learning algorithms are the
following: convergence is not always guaranteed, many algorithmic parameters are involved, sometimes long training times are required~\cite{Jaeger09,Doya92}.
For all those reasons learning using recurrent neural networks is principally feasible for relatively small networks.
Recently, a new paradigm called Reservoir Computing (RC) has been developed which overcome the main drawbacks of learning algorithms
applied to networks with cyclic topologies.
About ten years ago two main RC models were proposed: Echo State Networks (ESNs)~\cite{Jaeger01}
and Liquid State Machines (LSMs)~\cite{Maass02}. 
Both models describe the possibility of using recurrent neural networks \textit{without} adapting the weight connections involved in recurrences.
The network outputs are generated using very simple learning methods such as classification or regression models.
The RC approach have been  successfully applied in many machine learning tasks achieving goods results, specially in temporal learning tasks~\cite{Jaeger09, Maass02, Verstraeten07}. 

%We will concentrate on the latter.\\
%
%This approach is based on the empirical observation that under certain hypothesis, a learning process restricted to the output weights is often sufficient to obtain an excellent performance in many learning tasks.
%
% The internal recurrent structure in LSM model is generated with a type of spiking neurons called Leaky Integrate and Fire (LIF) neurons~\cite{Maass02}, while the the internal recurrent structure in the ESN model is built using analog sigmoidal functions.\\
%
%
%The reservoir in the LSM model is generated with a type of spiking neurons called Leaky Integrate and Fire (LIF) neurons~\cite{Maass02}, while the reservoir structure in the ESN model is built using analog sigmoidal functions~\cite{Jaeger09}.
%
In this paper we introduce a new type of RC method which uses some ideas
from RandNNs.
%
%Here we give a brief introduction of RandNN model, the basic concepts of RC methods and we study the building of reservoir with RandNNs.  We present some experimental results with this new model.
% Thus this paper we present a new type of RC method which is built using the RandNN model.....
%
%....
%and merges the properties innerents of RC models and RandNNs. 
%We study the performance of this new model, and the  using we present a..... as well as the performance results of the new model.......
%En este paper proponemos investigar el comportamiento de una RandNNs using the RC approach, that mean utilizar como internal neuronas RandNNs.  Para verificar el comportamiento heos probado con los siguetnes data set....\\
%
The paper is organized as follows: we begin by describing the RandNN model in Section~\ref{RandNNsmodel}.
In Section~III, we introduce the two funding RC models.
Section~\ref{ESQN} discusses the contribution of this paper,
a new RC model similar to the ESN, but using also
ideas inspired by queuing theory.
Finally, we present some experimental results and we end with some conclusions as well as a discussion regarding future lines of research.
%
%%%%%%

\section{Description of the \\
Random Neural Network Model}
\label{RandNNsmodel}

%Aclarar que se usa RandNNs para Random Neural networks y lo de los pesos su notacion.
%
A Random Neural Network (RandNN) is a specific queuing network proposed in~\cite{Gel89:RNNPosNeg} which merges concepts from spiking neural networks and queuing theory.
Depending on the context, its nodes are seen as 
queues or as spiking neurons.
Each of these neurons receives spikes (pulses) from outside, which are of one out of two disjoint types, called excitatory (or positive) and inhibitory (or negative).
Associated with a neuron there is an integer variable called the
neuron's potential. 
Each time a neuron receives an excitatory spike,
its potential is increased by one. If a neuron receives an inhibitory 
spike and its potential was strictly positive, it decreases by one;
if it was equal to~$0$, it remains at that value.
As far as the neuron's potential is strictly positive, the neuron
sends excitatory spikes to outside. When the neuron's potential is
strictly positive, we say that the neuron is excited or active.
After numbering the neurons in an arbitrary order,
let's denote by~$S_u(t)$ the potential of neuron~$u$ at time~$t$.
During the periods when the neuron is active,
it produces excitatory spikes with some rate~$r_u > 0$.
In other words, the output process of the pulses
coming out of an active neuron is a Poisson process. 
Then, a spike produced by neuron~$u$ is transferred to
the environment with probability~$d_u$.
For each synapse between neuron~$u$ and~$v$ an excitatory spike (respectively inhibitory spike) produced by~$u$ is  switched to neuron~$v$ with probability~$\ppuv$ (respectively~$\pnuv$). 
%%%
In the literature related to RandNNs, the probability that a pulse
generated at neuron~$u$ goes to neuron~$v$ is usually denoted by~$p_{u,v}^{+/-}$. 
This is different from the notation used in the standard ANNs literature, where a direct connection between~$u$ and~$v$ is often denoted as~$({v,u})$, that is, in the reverse order.
% denoted the probability of a pulse from neuron~$v$ vers neuron~$u$ occurs. 
In this paper we follow the latter notation.
This routing procedure is performed independently
of anything else happening in the network,
including previous or future switches
at the same neuron or at any other one. 
Observe that for any neuron~$u$ we have
   $$ d_{u} + \sum_{v=1}^{N} \ppuv + \sum_{v=1}^{N} \pnuv = 1,
   $$
where~$N$ is the number of neurons in the network. The weight connection between any two neurons~$u$ and~$v$ ($u$ sending spikes to $v$) is defined as:~$\wpuv=r_u\ppuv$ and~$\wnuv=r_u\pnuv$.

Let us assume that the external (i.e. from the environment)
arrival process of positive (respectively
negative) spikes to neuron~$u$ is Poisson with rate~$\lpu$ (respectively
with rate~$\lnnu$).
Some of these rates can be~0, meaning that no spike of the considered
type arrives at the given neuron coming from the network's environment.
In order to avoid the trivial case where nothing happens, we also must assume that~$\sum_{u=1}^{N} \lpu > 0$ (otherwise, the
network is composed of neurons that are inactive at all times). 
%
%The firing activity of different neurons at time~$t$ occurs independently between neurons and only depends on the internal state of the neuron itself. 
Last, the usual independence assumptions between all the considered Poisson and routing processes in the model are assumed.

We call~$\State(t) = (S_1(t),\cdots,S_N(t))$ the state of the network
at time~$t$. Observe that~$\State$ is a continuous time Markov process over the state space~$\mathbb{N}^N$. We will
assume that~$\State$ is irreducible and ergodic. We are interested in the network's behavior in steady-state, so, let
us assume that~$\State$ is in equilibrium (that is, assume~$\State$ is
stationary).
Let~$\loadu$ be the probability (in equilibrium) that neuron~$u$ is excited,
   $$ 
   \loadu = \lim_{t \rightarrow \infty} \Prob(S_u(t) > 0).
   $$
This parameter is called the \textit{activity rate} of neuron~$u$.
Since process~$S$ is ergodic, for all neron~$u$ we have $0 < \loadu < 1$.
Gelenbe in~\cite{Gel89:RNNPosNeg,Gel91:ProdFormQNet} shows that in an equilibrium situation the~$\loadu$s   
%stationary
%distribution of~$\bm{S}(t)$ can be written as the product of the marginal probabilities of the neuron's potential, that is:
%\beq
%\label{pf} \lim_{t \rightarrow \infty} \Prob(\State\,(t) =
%(n_1,\cdots,n_N)) = \prod_{i=1}^N (1 - \loadu) \loadu^{n_i}, 
%\eeq
%where the~$\loadu$s satisfies the following non-linear system of
satisfy the following non-linear system of
equations: 
	\begin{equation}\label{rhos} 
	\mbox{for each node~$u$, } \quad 
	\loadu = \frac{\Tpu}{r_u + \Tnu}, 
	\end{equation} 
	\begin{equation}\label{T+} 
	\mbox{for each node~$u$, }
		\quad \Tpu = \lpu + \sum_{v=1}^N \load_v \wpvu, 
	\end{equation} 
	\begin{equation}
	\label{T-}
	\mbox{for each node~$u$, }
		\quad \Tnu = \lnnu + \sum_{v=1}^N \load_v \wnvu,
	\end{equation}
with the supplementary condition that, for all neuron~$u$,
we have $\loadu < 1$.
In other words, under the assumption of irreducibility, if the system
of equations~(\ref{rhos}), (\ref{T+}) and (\ref{T-}) has a
solution~$(\load_{1},\cdots,\load_{N})$
such that we have~$\loadu < 1$, for all neuron~$u$, then the solution is unique
and the Markov process is ergodic. Moreover, its stationary distribution is given by the product of the marginal probabilities of the neuron's potential.
For more details and proofs, see~\cite{Gel89:RNNPosNeg,Gel98:LearningAndApprox}.
%%
%
%
%The model has been widely used in fields such as: machine learning problems, communication networks and computer systems.

As a learning tool used to learn some unknown function, we map
the function's variables to the external arrival rates,
the $\lpu$s and $\lnnu$s numbers (however, usually we set
set $\lnnu = 0$ for all input neuron~$u$, so we map the function's
variables to the $\lpu$s only).
The network's output is the set of loads.
The learning parameters are the set of weights in the model.
An appropriate optimization method (such as Gradient Descent) is used to find
weights such that when the arrival rate (of positive spikes) equals the input data,
the network output matches (with small error) the corresponding known output data
values.
The model has been widely used in fields such as: combinatorial optimization, machine learning problems, communication networks and computer systems~\cite{GelenbeXS99,Sakellari10,Rubi05:PSQA,Gelenbe1992,Cancela04}.
\section{Reservoir computing methods}
\label{RC}
Recurrent Neural Networks are a large class of computational models
used in several applications of Machine Learning and in neurosciences. 
The main characteristic of this type of ANNs is the existence of
at least one feedback loop among the connections, that is, a
circuit of connections.%
The cyclic topology causes that the non-linear transformation of the input history can be stored in internal states.
Hence recurrent neural networks are a powerful tool for forecasting
and time series processing applications. They are also very useful
for building associative memories, in data compression and for
static pattern classification~\cite{Jaeger09}.
%
%
%
%This form memory of past is sued to capture dynamical information
%
However, in spite of these important abilities and of the fact
that we have efficient algorithms for training neural networks without recurrences, no efficient algorithms exist for the case
where recurrences are present.

%
%
% almacened thehace que preserve en internal estados una transformacion no lineal de la entrada.
% the rnnn tenga importante capacidades para procesar temporal context information. capabilities to process temporal information come from their  
% th high power and accuracy is based on their cyclic topology which can be used to memory
%%%
Since the early 2000s, Reservoir Computing has gained prominence in the ANN community. In the two basic forms of the model described before,
ESNs and LSMs, at least three well-differenced structures can be identified:
the \textit{input layer}, where neurons receive information from the environment; the \textit{reservoir} (in ESNs) or \textit{liquid} (in LSMs), a nonlinear ``expansion'' function implemented using a recurrent neuronal network; the \textit{readout}, which is usually a linear function or a neural network without recurrences, producing the desired output.

The weight connections among neurons in the reservoir and the connections
between input and reservoir neurons are fixed during the learning process, only the weights between input neurons and readout units, and between reservoir and readout units, are the object of the training process.
The reservoir with its recurrences or circuits, allows
a kind of ``expansion'' of the input and possibly of
history data into a larger space. 
%
%A mapping from input data into a higher dimensional space, the original data inputs can gain meaningful linear structure in the space produced by the reservoir.
%
From this point of view, the reservoir idea is similar to the expansion function used in \textit{Kernel Methods}, for example
in the \textit{Support Vector Machine}~\cite{Vapnik95}.
The projection can enhance the linear separability of the data~\cite{Verstraeten07}.
On the other hand, the readout layer is built to be performant in learning, specially to be robust and fast in this process.
The RC approach is based on the empirical observation that under certain assumptions, training only a linear readout is often sufficient to achieve good performance in many learning tasks~\cite{Jaeger09}.
For instance, the ESN model has the best known learning performance on the Mackey--Glass times series prediction task~\cite{Schmidhuber07,Jaeger04}.
%
%That is based on the idea that a large pool of hidden neurons is capable to generate very rich dynamics and it is enough setting the output weights appropriately.
%

The topology of a RC model consists of an input layer with~$\Na$ units
sending pulses to the reservoir (and possibly also to the readout),
a recurrent neural network with~$\Nx$ units,
where $\Na \ll \Nx$, and a layer with~$\Nb$ readout neurons having
adjustable connections from the reservoir (and possibly
from the input) layer(s).
%
%
%

%\subsection{Liquid State Machines}
%
%The \textit{Liquid State Machines} (LSMs) is a computational model which has been proposed by W. Maass in~\cite{Maass02}. 
%
%The model is a mapping from continuous input and output streams in a multi-dimensional space and it comes from the interest to represent the cortical microstructures in the brain.
%
%The liquid is a structure which serves the readouts as an information pre-processor, while the readout neurons are fast information processors.
%
%The liquid requires to integrate all temporal information that is needed by the readout layer, due to the fact that the readout layer is a static maping without tmporal integration capabilities.
%
%This limitation of the readout comes from the motivation to produce  learning step as fast and robust as possible.
%
%Hence, the readout should be a function which can learn using simple algorithms.
%
%In the original LSM model the liquid was built using a model derived from Hodgkin-Huxley neuron model: Leaky Integrate and Fire (LFI) neurons. Other constructions have been suggested such as reservoir with dynamical synapses and threshold logic rates~\cite{Verstraeten07}.
%
%An important theoretical contribution of the LSM model is the analysis of the computational power of the model, in~\cite{Maass99} W. Maass  have shown that spiking neurons are computationally  more powerful than sigmoidal neurons. 
%

%\subsection{Echo State Networks}

The main difference between LSMs and ESNs consists in the type of nodes included in the reservoir. 
In the original LSM model the liquid was built using a model derived from Hodgkin-Huxley's work, called Leaky Integrate and Fire (LFI) neurons.
% Other constructions have been suggested such as reservoir with dynamical synapses and threshold logic rates~\cite{Verstraeten07}. 
%
In the standard ESN model, the activation function of the units is most often $\tanh(\cdot)$.
An ESN is basically a three-layered NN where only the hidden layer has recurrences, but allowing connections from input to readout (and, again,
where learning is concentrated in the readout only). 
Our training data consists of $K$ pairs $(\vak,\vbk)$, $k = 1,\ldots,K$, of input-output values of some unknown function $f$, where $\vak\in\R^{\Na}$,
$\vbk\in\R^{\Nb}$ and $\vbk = f(\vak)$.
The weights matrices are $\wi$ (connections from input to reservoir),
$\wre$ (connections inside the reservoir) and $\wo$ (connections between input or reservoir and readout), of dimensions
$\Nx\times(1+\Na)$, $\Nx\times\Nx$ and $\Nb\times (1+\Na+\Nx)$,
respectively. The first rows of $\wi$ and $\wo$ contain ones
corresponding to the bias terms.
\newcommand{\wmi}[1]{w^{\text{#1}}_{mi}}
%
%The dynamics of the reservoir are modeled at any discrete time $t>0$ by $\x(t)=\big(x_1(t),x_2(t),\ldots, x_{\Nx}(t) \big)$ where:
%%%
%\begin{equation}
%\tilde{x}_m(t)=\tanh\bigg( \ds{w_{m0}^{\text{in}} + \sum_{i=1}^{\Na} \wmi{in} a_i(t)} + \ds{\sum_{i=1}^{\Nx}\wmi{r} x_i(t-1)}\bigg),
%\end{equation}
%
%\begin{equation}
%\label{reservoirState}
%x_m(t)=\alpha \tilde{x}_m(t) + (1-\alpha)x_m(t-1), \forall m\in[1,\Nx],
%\end{equation}
%
%The network output  $\y=(y_1,\ldots, y_{\Nb})$ is calculated as follow:
%\begin{equation}
%\label{readout}
%y_m(t)=\ds{w_{m0}^{\text{out}} + \sum_{i=1}^{\Na} \wmi{out} a_i(t)} + \ds{\sum_{i=1}^{\Nx}\wmi{out} x_i(t)},  \forall m\in[1,\Nx],
%\end{equation}
%where $\alpha\in(0,1]$ is a parameter of memory control called \textit{leaking rate}. The initial state vector $\x(0)$ is arbitrarily chosen.

Each neuron $j$ of the reservoir has a real state $x_j$.
When the input $\va$ arrives to the ESN, the reservoir first
updates its state $\x = (x_{1}, \ldots,x_{N_{x}})$ by executing
%
%Usually is the ESN state is presented in vector notation
\begin{equation}
\label{reservoirStateVector}
{\x} := \tanh\big( \wi [1;\va] +\wre\x \big),
\end{equation}
and then, the ESN computes its outputs
\begin{equation}
\label{readoutVector}
\y := \wo [1;\va;\x],
\end{equation}
where $[\cdot;\cdot]$ is the vertical vector concatenation.

If we think of the ESN has a dynamical system receiving
a time series of inputs $\va(1), \va(2), \ldots$ and producing
a series of outputs $\y(1), \y(2), \ldots$, the corresponding series
of state values evolves according to
	$$ \x(t) = \tanh\big( \wi [1;\va(t)] +\wre\x(t-1) \big),
	$$
with the output at $t$ computed by $\y(t) := \wo [1;\va(t);\x(t)]$.

%An important principle which characterize the ESN model is that %
%Other weights are deemed fixed during the learning algorithm.
%

%A theoretical property is considered under the name of \textit{Echo State Property} (ESP)~\cite{Jaeger01} which informally described, 
% in the ESN case or presented as \textit{separability property} by Maass in LSM literature~\cite{Maass02}.
%
%Additionally, Maass prooved that LSM model have the \textit{universal computational power} property, when the liquid satisfies \textit{separability property} and the class of functions used to generate the readout satisfies \textit{approximation property}~\cite{Maass10}.
%
%We can informally describe this property as: the dynamics of the reservoir should converge given the same input, not taking into account the previous history~\cite{Mantas12}.
%
%In other words, a current output depends on inputs from a finite windows in the past~\cite{Verstraeten07}.
%
To ensure good properties in the reservoir, the $\wre$ matrix is
usually scaled to control its spectral radius
(to have $\rho(\wre)<1$)~\cite{Jaeger01}.
The role of the spectral radius is more complex
when the reservoir is built with spiking neurons
(in the LSM model)~\cite{Verstraeten07,Paugam09}.

%  and totally vanishes when an STDP rule is applied to the reservoir.
% A comparative study of several measures for the reservoir dynamics, with different neuron models can be found in....
%A structured algorithm to built reservoirs is still lacking.\\
%
%Finally, the reservoir weights are scaled by certain constant to ensure the \textit{Echo State Property} properties.
%
%
%
%
% Hablar de la continuidad de la liquid state networks
% De las propiedades .....
%El echo de streams hace que no necesite punto fijo en RNN, simplemente voy calculando en tiempos discretos.
%
%Aclarar bien las estabilidad de los nodos de RNN
%dynamical synapses and threshold logic rates~\cite{Verstraeten07}, 
Several extensions of the two pioneering RC models have been suggested in the literature, such as: intrinsic plasticity~\cite{Schrauwen07}, backpropagation-decorrelation~\cite{Steil04}, decoupled ESN~\cite{Xue07}, leaky integrator~\cite{Jaeger07}, Evolino~\cite{Schmidhuber07}, etc.
\section{A new Reservoir Computing method: \\
Echo State Queuing Networks}
\label{ESQN}
In this paper, we propose to reach the objective of simultaneously
keeping the good properties of the two models previously described.
%For this purpose, we propose the \textit{Echo State Queue Network} (ESQN), where we use RandNNs (that can be seen as a specific type of network of queues) in the reservoir of a ESN.
%
For this purpose, we introduce the \textit{Echo State Queuing Network}
(ESQN), a new RC model where the reservoir dynamics is based on a
specific type of queuing network (RandNN) behavior in steady-state.

%  (that can be seen as ) in the reservoir of a ESN.
%
%In~\cite{Gelenbe93}, E. Gelenbe suggested how to use RandNNs
%as a learning machine. The approach consists in a mapping of
%the rates of arrival signals into the activity rates ($\load$s)
%of the reservoir neurons,
%considering the dynamics of the network in steady-state.
%
%
%
%%\msb{This new model is a mapping between an input space in~$\R^{\Na}$ onto an output space in~$\R^{\Nb}$}. 
%%%
%%\msb{Both input and output data can be streams in discrete or continuous time.}
%
%
%in the spaces  and , respectively.
%
%

%
%In steady-state no tiene sentido usar $t$, pero aqui lo inluimos explicitamente porque se refiere a un indice correpondiente al stream de entrada de la maquina.

%We denote the set of input neurons by~$\mathcal{I}$ and the set of neurons in the reservoir by~$\mathcal{R}$.\\
%

The architecture of an ESQN consists of an input layer, a reservoir and a readout layer.
The input layer is composed of~$\Na$ random neural units which send spikes toward
the reservoir or toward the readout nodes.
The reservoir dynamics is designed inspired by the equations
of recurrent RandNNs (see below).
Let us index the input neurons from~$1$ and~$\Na$, and the reservoir
neurons from~$\Na+1$ to~$\Na+\Nx$.
%
% As presented in Section~\ref{RandNNsmodel}, the spikes among random neural neurons can be positive or negative and are characterized by the mean rates of emission.  For any two neurons~$u$ and~$v$ the mean rate of spikes sent to neuron~$v$ from~$u$ is

When an input $\va$ is offered to the network, we first
identify the rates of the external positive spikes
with that input, that is:~$\lpu = a_u$, and, as it is
traditionally done in RandNNs,
$\lnnu = 0$, for all $u = 1,\ldots,\Na$. 
In a standard RandNN, the neuron's loads are computed solving the
expressions~(\ref{rhos}),~(\ref{T+}) and~(\ref{T-}). More precisely,
input neurons behave as a $M/M/1$ queues. The load or activity
rate of neuron~$u$, $u = 1,\ldots,N_{a}$ is, in the stable case
($a_u < r_{u}$), simply $\loadu = a_u/r_{u}$. 
For reservoir units, the loads are computed solving the non-linear system
composed of equations~(\ref{rhos}), (\ref{T+}) and (\ref{T-}).
The network is stable if all obtained loads are $< 1$.

% $u$, $u = N_{a}+1,\ldots,N_{a}+N_{x}$, the corresponding load
%is
%	$$
%		\loadu = \ds{\frac{\ds{\sum_{v=1}^{\Na+\Nx}\load_v \wpuv}}%
%					{r_u+ \ds{\sum_{v=1}^{\Na+\Nx}\load_v\wnuv}}},
%	$$
%when this non-linear system has a solution where all the loads are $< 1$.

In our ESQN model, we do the same for input neurons, but for the reservoir,
we introduce the concept of state. The state is simply the
vector of loads $\mathbf{\varrho}$. When we need the network output corresponding to a
new input $\va$, we first compute a new state by using
\begin{equation}
\label{reservoirState}
	\loadu := \ds{\frac{\ds{\sum_{v=1}^{\Na}\frac{a_v}{r_v} \wpvu + \sum_{v=\Na+1}^{\Na+\Nx}\load_v \wpvu}}%
		  {r_u+ \ds{\sum_{v=1}^{\Na}\frac{a_v}{r_v} \wnvu + \sum_{v=\Na+1}^{\Na+\Nx}\load_v \wnvu}}},
\end{equation}
for all $u\in [\Na+1,\Na+\Nx].$  
%
%%
%
%\begin{equation}
%\label{reservoirState}
%	\loadu := \left[ \sum_{v=1}^{\Na+\Nx}\load_v \wpuv \right]% 
%		\!\!	\left[ r_u+ \ds{\sum_{v=1}^{\Na+\Nx}\load_v\wnuv} \right]^{\!\! -1}.
%\end{equation}
%
%\begin{equation}
%\label{reservoirState}
%	\loadu := \ds{\frac{\ds{\sum_{v=1}^{\Na+\Nx}\load_v \wpuv}}%
%					  {r_u+ \ds{\sum_{v=1}^{\Na+\Nx}\load_v\wnuv}}}.
%\end{equation}
When this is seen as a dynamical system, on the left we have the
loads at $t$, and on the r.h.s. the loads at $t-1$.

The readout part is computed by a parametric function~$g(\wo,\va,\mathbf{\varrho})$
(or $g(\wo,\va(t),\mathbf{\varrho}(t))$ when this is used as a dynamical prediction system.
%which receives the reservoir state~$\x(t)$ and produces an output~$\y(t)\in\R^{\Nb}$}.
%
In this paper we present the simple case of computing the readout using a
linear regression. 
It is easy to change this by another type of function $g$,
due to the independent structure between the reservoir and readout layers.
Thus, the network output~$\y(t)=[y_1(t),\ldots, y_{\Nb}(t)]$ is computed for any~$m\in[1,\Nb]$ using expression~(\ref{readoutVector}) and it can be written as follows, where we use the temporal version at time~$t$:
\begin{equation}\label{readout}
%y_m(t)=\ds{w_{m0}^{\text{out}} + \sum_{i=1}^{\Na} \wmi{out} \lambda^+_i(t)} + \ds{\sum_{i=1+\Na}^{\Na+\Nx}\wmi{out} \load_i(t)}.
	y_m(t) = \ds{w_{m0}^{\text{out}} + \sum_{i=1}^{\Na} \wmi{out} a_i(t)}
				+ \ds{\sum_{i=1+\Na}^{\Na+\Nx}\wmi{out} \load_i(t)}.
\end{equation}
The output weights~$\wo$ can be computed using some of the traditional algorithms to solve regressions such as the ``ridge regression'' or the least mean square algorithms~\cite{numericalRecips92}.

\section{Experimental Results}

In our numerical experiences, we consider a simulated time series
data widely used in the ESN literature~\cite{Jaeger01,Rodan11}
and two real world data sets about Internet traffic, used in research work about forecasting techniques~\cite{Cortez12,timesSeries}.
To evaluate the models' accuracy, we use
the \textit{Normalized Mean Square Error} (NMSE):
%, due to the fact that our results can be compared with another ones presented in the literature.
%NMSE is defined as:
%\begin{equation}
%\label{MSE}
%MSE=\displaystyle{\frac{1}{T} \sum_{t=1}^{T}\sum_{j=1}^{\Nb}{\big(y_{j}(t)-{b_{j}(t)\big)^2}}},
%\end{equation}
%
\begin{equation}
\label{MSE}
	\text{NMSE} = \ds{\frac{\sum_{k=1}^{K}\sum_{j=1}^{\Nb}
					{\big({b_{j}^{(k)}-y_{j}^{(k)}\big)^2}}}%
		{\sum_{k=1}^{K}\sum_{j=1}^{\Nb}
			{\big({b_{j}^{(k)}-\bar{b}_{j}\big)^2}}}},
\end{equation}
where~$\bar{b}_{j}$ is the empirical mean, and where we use the
same notation as before for the data. 
The positive and negative weights of the ESQN model and the initial reservoir state were randomly initialized in the intervals~$[0,0.2]$ and~$[0,1]$, respectively.
As usual, the training performance can depend on the choice of the starting weights. To take this into account, we experiment with~$20$ different random initial weights and we calculate their average performance.
%
% the input and reservoir weights~$\{\wpuv, \wnuv \forall u,v=1,\ldots,\Na+\Nb\}$ and the readout weights~$\w^{\rm{out}}$. T.
%
The preprocessing data step consisted in rescaling the data in the interval~$[0,1]$.
The learning method used was offline ridge regression~\cite{Mantas12}. This algorithm contains a regularization parameter which is adjusted for each data set.
%
% equal to~$0.1$.
%
The time series data considered were: 
% We consider the following time series data:
%
\begin{enumerate}
%
%\item H\'enon Map dataset~\cite{Rodan11}: the length of training is~$4000$ and~$800$ the size of validation data, is generated by:
%$y_{\rm{target}}(t)=1-1.4y_{\rm{target}}(t-1)^2+0.3y_{\rm{target}}(t-2)+z(t)$,
%where the noise is~$z(t)\sim\mathcal{N}(0,0.05)$. The serie is shifted by~$-0.5$ and scaled by~$2$ as it is recommended in~\cite{Rodan11}.
%%
\item Fixed~$10$th order nonlinear autoregressive moving average (NARMA) system.
The series is generated by the following expression:\\
$b(t+1)=0.3\,b(t)+0.05\,b(t){\sum_{i=0}^{9}b(t-i)}\\
+1.5\,s(t-9)\,s(t)+0.1,$
where~$s(t)\sim \text{Unif}[0,0.5]$. We generated a training data with~$1990$ samples and a validation set with~$390$ samples. 
\item Traffic data from an Internet Service Provider (ISP)
working in~$11$ European cities.
% A traffic forecasting study of this data was  presented in~\cite{Cortez12}. 
The original data is in bits and was collected every~$5$ minutes. We rescaled it in~$[0,1]$.
%
%We present the results for the time series when the original data is summed within intervals of~$5$ minutes. 
%
The size of the training data is~$9848$ and the size of validation set is~$4924$.
The input neurons ($\Na=7$) are mapped to the last~$7$ points of the past data, that is with values from~$t-6$ up to time~$t$. This configuration was  suggested in~\cite{Cortez12} where the authors discuss different neural network topologies taking into account
seasonal traits of the data.
\item Traffic data from United Kingdom Education and Research Networking Association (UKERNA). 
The Internet traffic was collected every day.
The network input at any time~$t$ is the triple composed
of the traffic at times $t$, $t-6$ and $t-7$, as studied
in~\cite{Cortez12}.
This small data set has $47$ training pairs and $15$ validation samples.
\end{enumerate}
\begin{table}[!t]
% \extrarowheight as needed to properly center the text within the cells
\caption{Comparison between ESN and ESQN.
We give the NMSE obtained from~$20$ independent trials, and the
corresponding Confidence Interval (CI), for an ESN model and
the proposed ESQN procedure. 
The reservoir size was~$80$ units for the first data set and~$40$~units for the other ones.
}
\label{tablePerformance}
\begin{center}
\begin{tabular}{|c|c|c|c|}
\hline
Series & Model & NMSE & CI\\
\hline
\multirow{2}{*}{NARMA} & ESN & $0.1401$ &  $\pm 0.0504$\\ %Nx=80, sparsity=0.15; alpha=0.95; rho=0.9; gamma=0.03;
$\;$ & ESQN  & $0.1004$  & $\pm  0.0025$ \\ % Nx= 80   
\hline
%European & NN & $$ & $$\\
\multirow{2}{*}{ISP} &  ESN  & $0.0062$ & $\pm 9.8885 \times 10^{-7} $\\ %gamma=0.1 40 units y rho=(0.5)
%%%
$\;$ & ESQN  & $0.0100$ & $\pm 1.2436  \times 10^{-4}$\\ 
%MSE 3.7050e-04, MAPE 2.2286,   var 8.2856e-09,  Nx 40. 30 runs, gamma=0.1, Winit=0.2,density=1,
\hline
%$\;$ & NN & $$ &  \\
% Corri 30 veces y selecciono los 20 mejores, descartando casos de inestabilidad.
%
\multirow{2}{*}{UKERNA} & ESN &  $0.3781$ & $\pm0.0066$\\ %  La densidad de los pesos es 0.2,alpha= 1, Selecciono los 20 mejores casos y los promedio.
$\;$ & ESQN  & $0.2030$  & $\pm0.0335$\\ % La densidad de los pesos es 0.4, cuando era 1 no funca. MSE=0.0168;  MSE alcanza a ser  0.0036 en el mejor caso.
% NMSE_ESQN alcanza a ser 0.0855. Nx es con i=2.
\hline
\end{tabular}
\end{center}
\end{table}
The NARMA series data was studied in deep in~\cite{Verstraeten07,Jaeger04,Rodan11,Atiya00,BasterCord11}.
For the last two data sets the performance using NN, ARIMA and Holt-Winters methods can be seen in~\cite{Cortez12}.
%
%Several papers investigates the ideas to built ``good'' reservoirs, however there are not still recipes to achieve them~\cite{Jaeger09, Rodan11}.
%
A typical ESN model consists in a reservoir with the following characteristics: random topology,~$\Nx$ large enough, sparsely connected (roughly between~$15\%$ and~$20\%$ of their weights are non-zeros)~\cite{Jaeger09}.
%
%Two sensible parameters in the ESN model are the sparsity and the spectral radius  of the reservoir weight matrix. 
%
The specific ESN used has a sparsity of~$15\%$ and spectral radius of~$0.95$ in its reservoir matrix. 
In~\cite{Verstraeten07}, the authors obtained the best performance in the NARMA data problem when the spectral radius was close to~$1$.

%In the standard ESN only between~$10\%$ and~$20\%$ of the reservoir weights are non-zero. 
%
The ESN performance can be improved using leaky-integrator neurons~\cite{Jaeger07}, feedback connections~\cite{Jaeger09} or initializing the reservoir weights using another initializing criteria~\cite{BasterCord11,Luko10}.
Both models have~$80$ units in the reservoir for the NARMA data and~$40$ units for the other two data sets.
In this paper, in order to compare the performance of the ESQN and ESN models we use the standard versions of each of them.

Table~\ref{tablePerformance} presents the accuracy of the
ESQN and ESN models. In the last column we give a~$95\%$
confidence interval obtained from $20$ independent runs.
We can see that for the $10$th order NARMA and UKERNA data the performance obtained with ESQN is better than with ESN
(even if in the NARMA case, the confidence intervals have a ``slight''
non-empty intersection).
In the case of the European ISP data, ESN shows a significant
better performance.
Observe that in all cases the accuracy obtained with ESQN was very good.
Also observe that we are using some years of cumulated knowledge
about ESNs in our implementation, which we are comparing with our first
versions of our new largely unexplored ESQN model.

\begin{figure}[!t]
\centering
\includegraphics[angle=0,height=2.5in,width=3.5in]{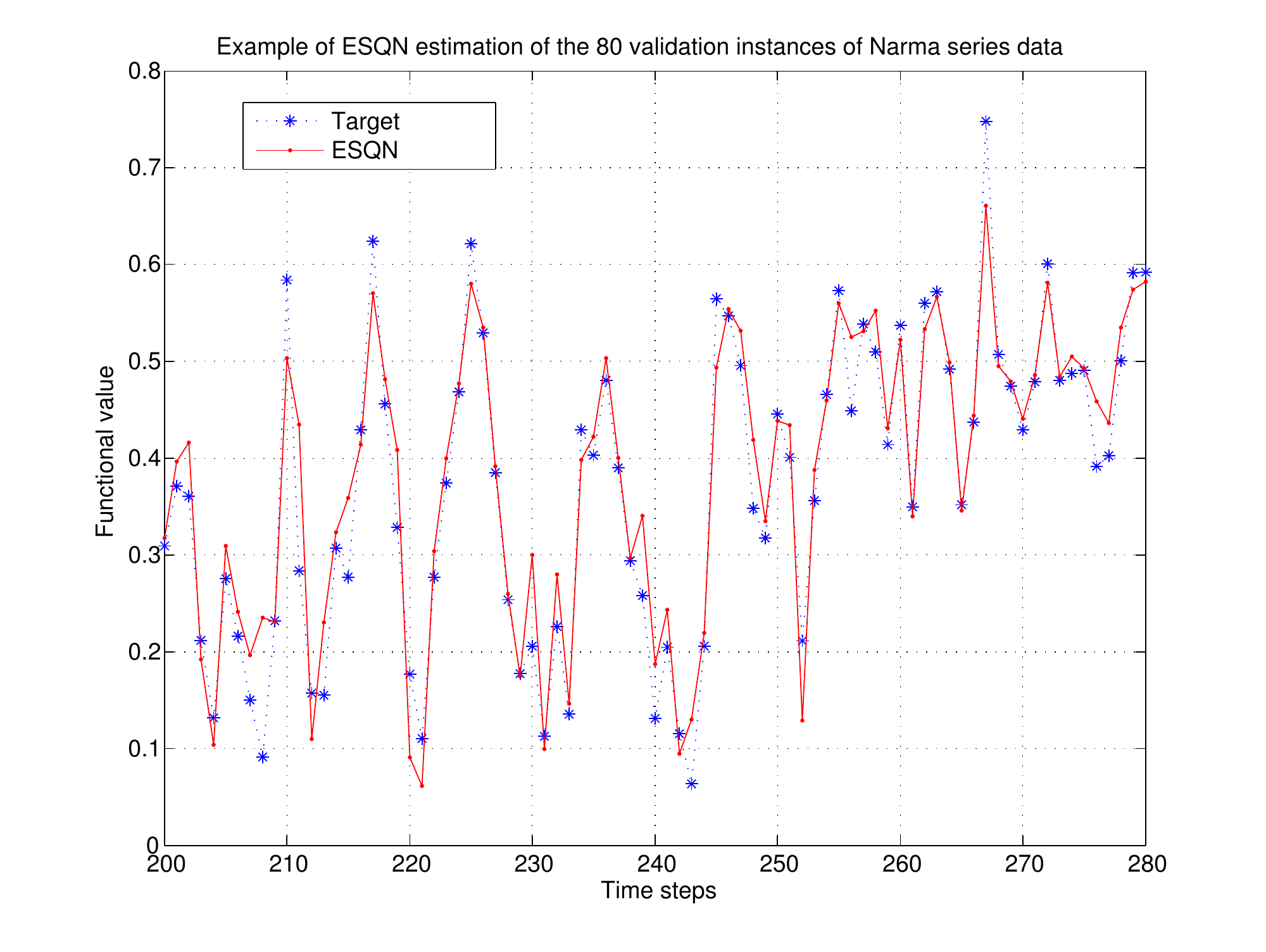}
\caption{Example of ESQN prediction for~$80$ time steps of fixed~$10$th NARMA validation data set. The reservoir was randomly initialized and it had~$80$ units.}
\label{NarmaEx}
\end{figure}

%
%
%\begin{figure}[!t]
%\centering
%\includegraphics[angle=0,height=2.5in,width=3.5in]{./Figures/ZoomA5M_1_250.eps}
%\caption{Example of ESQN prediction for the first~$250$ instances in the validation set of the European ISP traffic data.
%
% The architecture of ESQN is~$6$ input neurons,~$40$ neurons in the reservoir and~$1$ output neuron. The reservoir was random initialized.}
%\label{A5MEx}
%\end{figure}
%
\begin{figure}[!t]
\centering
\includegraphics[angle=0,height=2.5in,width=3.5in]{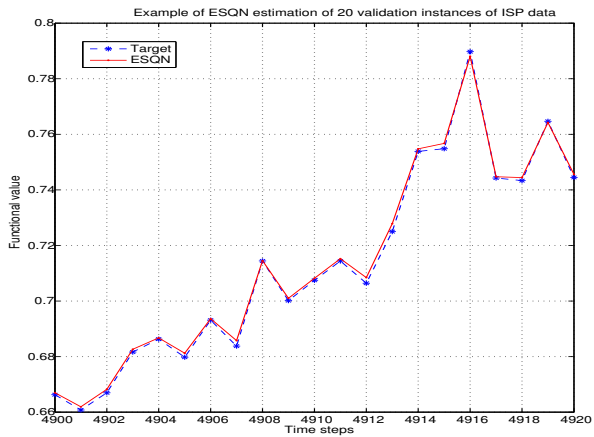}
\caption{Example of ESQN prediction for~$20$ instances in the validation set of the European ISP traffic data. The instances correspond to time steps between~$4900$ and~$4920$.
The reservoir was randomly initialized and it had~$40$ neurons.}
\label{A5MEx2}
\end{figure}
\begin{figure}[!t]
\centering
\includegraphics[angle=0,height=2.5in,width=3.5in]{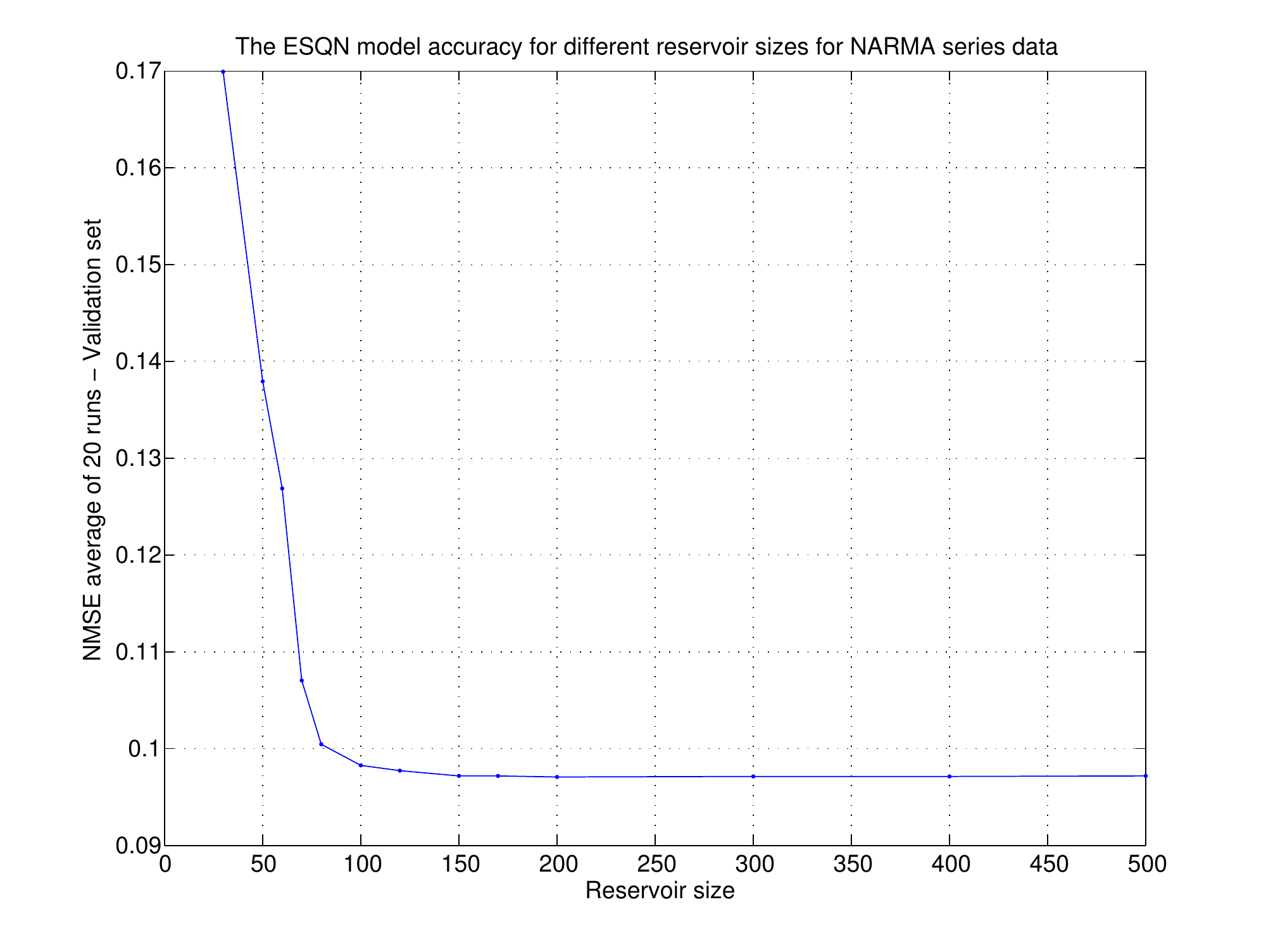}
\caption{The ESQN model performance for different reservoir sizes which are computed for~$10$th NARMA validation data set. The reservoir weights were randomly initialized. Figure shows the NMSE average achieved in~$20$ runs with different ESQN initial weights.}
\label{NarmaVsReservoirSize}
\end{figure}
%
%
%\begin{figure}[!t]
%\centering
%\includegraphics[angle=0,height=2.5in,width=3.5in]{./Figures/EQNESN_UKERNA1D_Initialization.eps}
%\caption{EQN and ESN model performance for different weight initialization for UKERNA validation data set. In both cases reservoir sizes had  40 units and were randomly initialized.}
%\label{EQNESN_UKERNA1D_Initialization}
%\end{figure}
%
%
\begin{figure}[!t]
\centering
\includegraphics[angle=0,height=2.5in,width=3.5in]{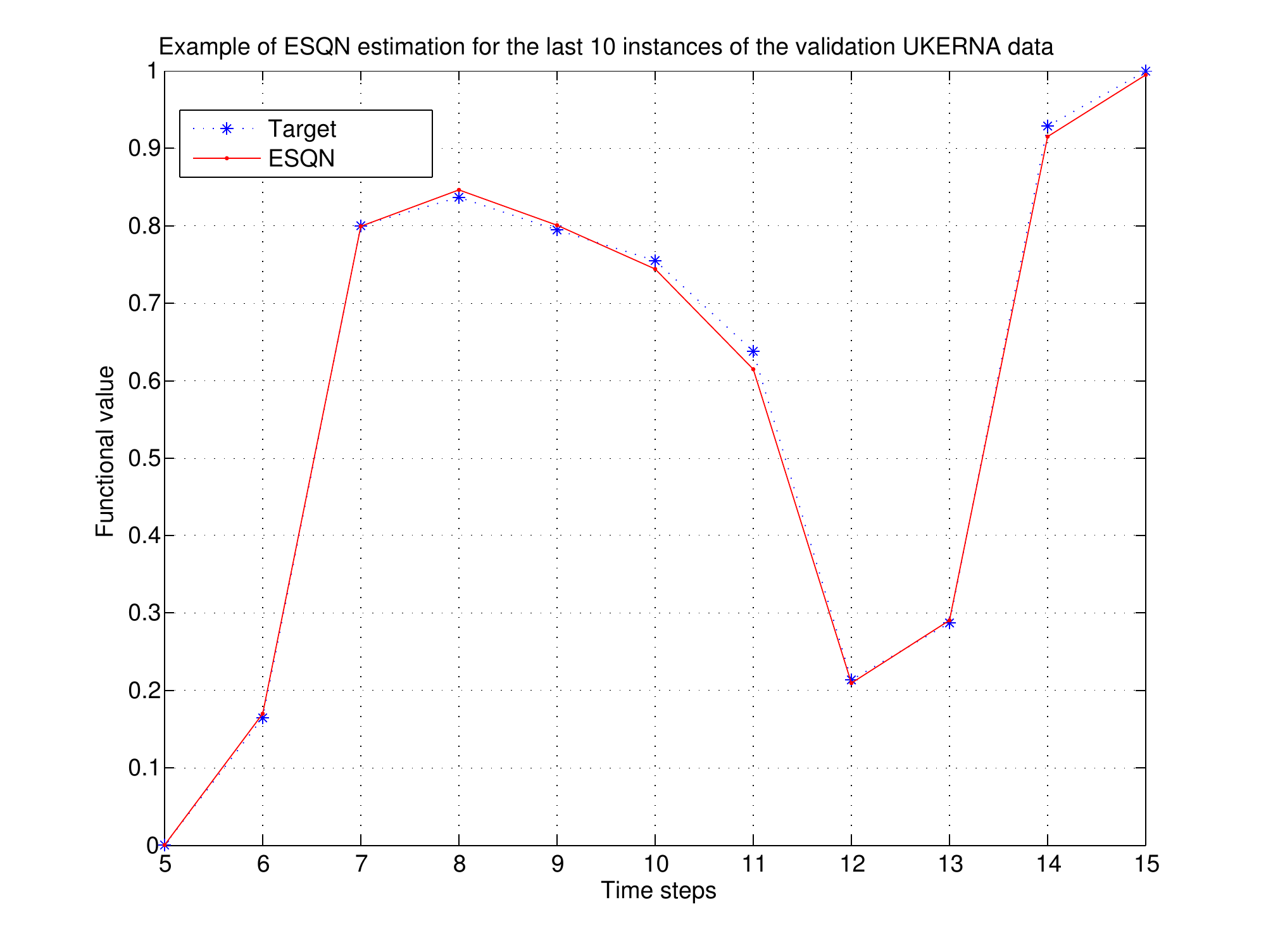}
\caption{ESQN estimation for UKERNA validation data set. The reservoir weights were randomly initialized. The reservoir size is~$40$.}
\label{validationUKERNA1D_EQN}
\end{figure}

%MWe show in the Figures~(\ref{Freedman},\ref{Henon},\ref{Narma10},\ref{Obser}) the performances of RAND, SIM and SOM. In all those cases the performances are better in the cases of SIM and SOM than the RAND case. There is important significant difference of performance in the case of H\'enon Map. On the other hand, in the other examples, Freedam dataset, Nonlinear System with Observational noise and~$10$th order Narma dataset, when the network is large, the SOM and SIM not improve significantly on the random reservoir. The Table~(\ref{Sd}) shows the standard deviation of the 50 different initial weights of ESN for each method and for each one of data set when the reservoir size is $\Nx=40$. 
%
%

Figure~\ref{NarmaVsReservoirSize} shows that the reservoir size is an important parameter affecting the performance of the ESQN. 
This also happens with the ESN model: in general,
a larger reservoir enriches the learning abilities of the model.
The sparsity and density of the reservoir in the ESQN model was not studied in this work. It is left for future efforts.
NARMA is an interesting time series data where the outputs depend on both the input and previous outputs. The modeling problem is difficult to solve due to the non-linearity of the data and the necessity of having
some kind of long memory.
Figure~\ref{NarmaEx} illustrates an estimation of ESQN with~$80$ units in the reservoir which are randomly chosen in~$[0,0.2]$.
%
%Decir que en el paper de Cortez se puede ver los resultados con NN arima y otros metodos.
%
Figures~\ref{A5MEx2} and~\ref{validationUKERNA1D_EQN} show
the prediction values for an interval of validation data. 
Figure~\ref{A5MEx2} shows the prediction of~$20$ instances beginning at time~$4900$ of the validation set.
The main difficulty to model UKERNA data (using day scale) is that the training set is small.
In spite of this, Figure~\ref{validationUKERNA1D_EQN} illustrates the good performance of the ESQN model. This figure shows the estimation of the last~$10$ instances in the validation data.
%
%
%Aclarar el algo usado para calcular los wout
%
%Estudiar en futuros papers cual es la complexity minima del reservoir  necesario para ....
% Destacar la facilidad de implementacion pues es un simple contador.
%\newpage
\section{Conclusions}

In this contribution, we have presented a new type of Reservoir Computing
model which we call Echo State Queuing Network (ESQN).
It combines ideas from queueing and neural networks.
It is based on two computational models: the Echo State Network (ESN)
and the Random Neural Network.
Both methods have been successfully used in forecasting and machine
learning problems. 
Particularly, ESNs have been applied in many temporal
learning tasks.
Our model was used to predict three time series data which are widely
used in the machine learning literature.
In all cases tested, the performance results have been very good.
%
%As other RC methods ESQN have shown  efficiency and 
%
We empirically investigated the relation between the reservoir size
and the ESQN performance.
We found that the reservoir size has a significant impact on the accuracy.
Another positive property of ESQNs is their simplicity,
since reservoir units are just counter functions.
Last, our tool is very easy to implement, both in software
and in hardware.

There are still several aspects of the model to be studied in future work.
Some examples are
the impact of the sparsity of the reservoir weights, the weight initialization methods used, the scaling of reservoir weights and the utilization of leaky integrators.
%
%As consequence, we believe that the ESQN model can be an interesting alternative to be applied in networking and machine learning problems.
\bibliographystyle{IEEEtran}
\bibliography{refRnn}
%\begin{thebibliography}{1}

%H.~Kopka and P.~W. Daly, \emph{A Guide to \LaTeX}, 3rd~ed.\hskip 1em plus 0.5em minus 0.4em\relax Harlow, England: Addison-Wesley, 1999.
% \end{thebibliography}

\end{document}